\documentclass[conference]{IEEEtran}
\IEEEoverridecommandlockouts
\usepackage{cite}
\usepackage{amsmath,amssymb,amsfonts}
\usepackage{algorithmic}
\usepackage{textcomp}
\usepackage{graphicx}
\usepackage{xcolor}
\usepackage{algorithm}
\usepackage{pifont}
\newcommand{\cmark}{\ding{51}} 
\newcommand{\xmark}{\ding{55}} 
\usepackage[caption=false,font=footnotesize]{subfig}
\def\BibTeX{{\rm B\kern-.05em{\sc i\kern-.025em b}\kern-.08em
    T\kern-.1667em\lower.7ex\hbox{E}\kern-.125emX}}
\begin{document}

\title{ReGAIN: Retrieval-Grounded AI Framework for Network Traffic Analysis

\thanks{Accepted to the 2026 IEEE International Conference on Computing, Networking and Communications (ICNC 2026). The final published version will appear in IEEE Xplore (DOI to be added when available).

© 2026 IEEE.  Personal use of this material is permitted.  Permission from IEEE must be obtained for all other uses, in any current or future media, including reprinting/republishing this material for advertising or promotional purposes, creating new collective works, for resale or redistribution to servers or lists, or reuse of any copyrighted component of this work in other works

Code, data, and model availability: https://github.com/270771/llm-traffic}
}

\author{\IEEEauthorblockN{Shaghayegh Shajarian}
\IEEEauthorblockA{\textit{Computer Science} \\
\textit{North Carolina A\&T State University}\\
Greensboro NC, USA \\
0009-0003-7334-3864}
\and
\IEEEauthorblockN{Kennedy Marsh}
\IEEEauthorblockA{\textit{Computer Science} \\
\textit{North Carolina A\&T State University}\\
Greensboro NC, USA \\
0000-0002-5987-252X}
\and
\IEEEauthorblockN{James Benson}
\IEEEauthorblockA{\textit{Institute for Cyber Security} \\
\textit{University of Texas at San Antonio}\\
San Antonio TX, USA \\
0000-0001-7209-2344}
\and
\IEEEauthorblockN{Sajad Khorsandroo}
\IEEEauthorblockA{\textit{Computer Science} \\
\textit{North Carolina A\&T State University}\\
Greensboro NC, USA \\
0000-0003-0649-9247}
\and
\IEEEauthorblockN{Mahmoud Abdelsalam}
\IEEEauthorblockA{\textit{Computer Science} \\
\textit{North Carolina A\&T State University}\\
Greensboro NC, USA \\
0000-0001-5627-5239}
}

\maketitle

\begin{abstract}

Modern networks generate vast, heterogeneous traffic that must be continuously analyzed for security and performance. Traditional network traffic analysis systems, whether rule-based or machine learning–driven, often suffer from high false positives and lack interpretability, limiting analyst trust. In this paper, we present ReGAIN, a multi-stage framework that combines traffic summarization,  retrieval-augmented generation (RAG), and Large Language Model (LLM) reasoning for transparent and accurate network traffic analysis. ReGAIN creates natural-language summaries from network traffic, embeds them into a multi-collection vector database, and utilizes a hierarchical retrieval pipeline to ground LLM responses with evidence citations. The pipeline features metadata-based filtering, MMR sampling, a two-stage cross-encoder reranking mechanism, and an abstention mechanism to reduce hallucinations and ensure grounded reasoning. Evaluated on ICMP ping flood and TCP SYN flood traces from the real-world traffic dataset, it demonstrates robust performance, achieving accuracy between 95.95\% and 98.82\% across different attack types and evaluation benchmarks. These results are validated against two complementary sources: dataset ground truth and human expert assessments. ReGAIN also outperforms rule-based, classical ML, and deep learning baselines while providing unique explainability through trustworthy, verifiable responses.

\end{abstract}


\section{Introduction} \label{intro}

Modern networks generate massive volumes of traffic that must be continuously monitored for performance, reliability, and security. Alongside real-time monitoring, historical traffic data (e.g, packet captures (PCAPs) and flow records) is invaluable for forensic investigations. These records often contain the clearest evidence of malicious activity, revealing abnormal payloads, tunneling techniques, or credential theft. Retrospective traffic analysis thus plays a critical role in reconstructing attack campaigns, validating alerts, and improving defensive strategies. However, traditional network traffic analysis systems face several limitations. Rule-based systems (e.g., Snort, Suricata) rely on manually crafted signatures, which require constant maintenance, produce high false positive rates, and offer limited explanations. Machine learning approaches, such as Support Vector Machines (SVMs), Random Forests, and deep learning models, achieve strong detection accuracy but often operate as black boxes. This lack of explainability reduces analyst trust and complicates incident response, as operators must manually correlate alerts with supporting evidence from multiple data sources. Large Language Models (LLMs) have shown promise in network operations \cite{asurvey,self-running}, with reasoning over semi-structured data, and generating human-readable insights. However, in purely generative modes, they risk hallucinations and unverifiable claims. Retrieval-Augmented Generation (RAG) mitigates this by grounding LLM outputs in external knowledge sources, ensuring that generated explanations are supported by verifiable evidence \cite{rag}.

In this paper, we present ReGAIN (Retrieval-Grounded AI for Network Traffic Analysis), a multi-stage, LLM-driven framework that integrates hierarchical semantic retrieval, evidence quality monitoring, and citation-backed reasoning. Building on our previous work \cite{shajarian2024}, ReGAIN comprises four primary components (detailed in Section~\ref{proposed-framework}): (1) a data ingestion pipeline that transforms heterogeneous network telemetry into natural language summaries, (2) a multi-collection vector knowledge base that semantically indexes these summaries with structured metadata for efficient retrieval, (3) a retrieval-augmented reasoning engine that combines multiple techniques (e.g., metadata filtering, cross-encoder reranking) to select high-quality evidence, and (4) an LLM-driven analysis component that generates human readable explanations with explicit citations to supporting records. Unlike traditional RAG systems that rely on a single knowledge base and lack retrieval quality controls, ReGAIN leverages multi-collection retrieval across specialized knowledge bases, adaptive context selection via automated metadata filtering and Maximal Marginal Relevance (MMR) diversity sampling, multi-stage retrieval refinement using bi-encoder search followed by cross-encoder reranking, and an abstention mechanism that returns diagnostic feedback when retrieval quality is insufficient, preventing hallucinations while providing human readable explanations with explicit citations to supporting records. The contributions of this work are as follows:

\begin{itemize}
    \item We introduce the ReGAIN framework, which unifies structured traffic representation, semantic embedding, vector retrieval, and LLM reasoning for network traffic analysis.
    \item We design a pipeline with multi-collection retrieval, adaptive context selection, multi-stage reranking, and an abstention mechanism that mitigates hallucinations.
    \item We conduct a dual-mode evaluation, combining automated and expert-based approaches, on a real-world network traffic dataset that includes ICMP ping flood and TCP SYN flood attacks, illustrating strong detection performance across both scenarios.
    \item We benchmark ReGAIN against traditional rule-based, classical machine learning, and deep learning approaches, demonstrating superior performance.
\end{itemize}

\section{Related Work} \label{related-work}

Recent studies have explored how LLMs can support networking tasks beyond traditional classifiers. NetLLM~\cite{netllm} adapts LLMs for a range of networking problems and represents an early step toward unified LLM-driven workflows in this area. ShieldGPT~\cite{shieldgpt} applies an LLM-driven approach to detect and mitigate DDoS attacks, showing how language-based reasoning can complement existing traffic analysis tools. In network security, several works emphasize that LLMs are most useful for generating explanations that help operators understand alerts rather than replacing detection systems completely. Houssel \emph{et~al.} evaluate LLMs as explainable components for intrusion detection and later propose eX-NIDS, which focuses on improving interpretability for flow-based NIDSs~\cite{houssel2024,houssel2025}. TrafficLLM~\cite{trafficllm} introduces domain-specific traffic representations and dual-stage fine-tuning to improve generalization across different network traffic tasks.

RAG has also started to gain attention in cybersecurity. Rahman \emph{et~al.} show that combining knowledge graphs with RAG can improve cyber threat analysis by linking model outputs to structured information. Other works utilize retrieval-enhanced LLMs to support incident response and decision making~\cite{pnnlrag,irllmrag}. Meanwhile, the broader security community has begun benchmarking and evaluating LLMs for security related applications, highlighting the need for domain-specific evaluation frameworks~\cite{ndss25-llm-eval}. Despite these advances, most existing LLM-driven systems focus on either detection accuracy or interpretability in isolation. To fill in this gap, ReGAIN contains retrieval-augmented reasoning that grounds LLM outputs in cited traffic evidence and related artifacts, a deterministic summarization layer that converts raw network logs into concise natural language descriptions for embedding, a hybrid evaluation strategy that combines ground truth comparison with human expert labels (see Table~\ref{tab:novelty-compare}).

\begin{table}[t]
\centering
\caption{Comparison of ReGAIN with related LLM-based systems.}
\label{tab:novelty-compare}
\setlength{\tabcolsep}{1.5pt}
\footnotesize
\begin{tabular}{lcccc}
\hline
\textbf{Framework} & \textbf{RAG} & \textbf{Traffic Repr.} & \textbf{Expert Val.} & \textbf{Evidence Cited} \\
\hline
NetLLM~\cite{netllm} & \xmark & Structured features & \xmark & Limited \\[-2pt]
 &  & (task adapters) &  &  \\[1pt]
ShieldGPT~\cite{shieldgpt} & \xmark & Flow-level inputs & \xmark & Generative \\[-2pt]
 &  &  &  &  \\[1pt]
eX-NIDS~\cite{houssel2025} & \xmark & Flow inputs + templates & \xmark & Template \\[-2pt]
 &  &  &  &  \\[1pt]
TrafficLLM~\cite{trafficllm} & \xmark & Generic traffic repr. & \xmark & Domain prompts \\[-2pt]
 &  &  &  &  \\[1pt]
\textbf{ReGAIN (ours)} & \textbf{\cmark} & \textbf{NL summaries + embed.} & \textbf{\cmark} & \textbf{Cited evidence} \\[-2pt]
 &  &  &  &  \\ \hline
\end{tabular}
\end{table}

\section{ReGAIN Framework} \label{proposed-framework}
\begin{figure*}[htbp]
\centerline{\includegraphics{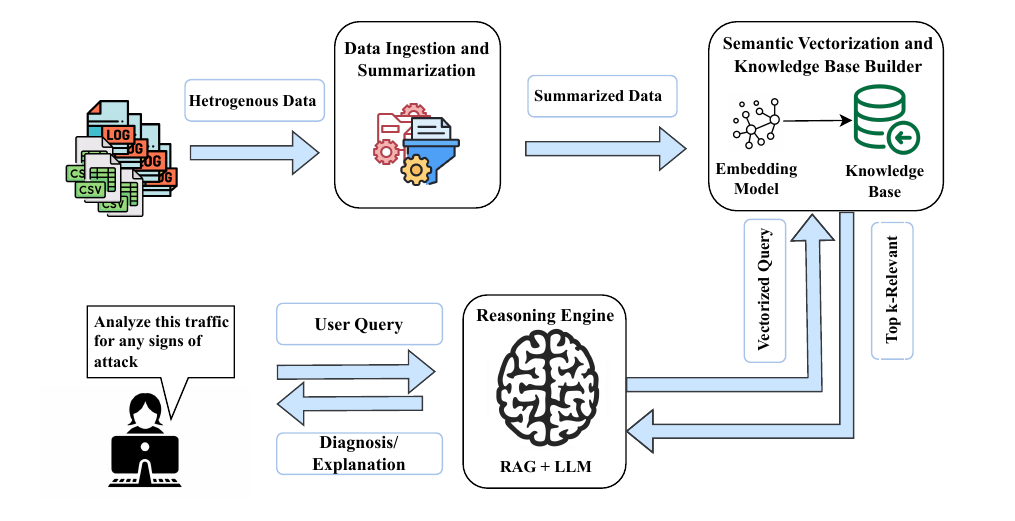}}
\caption{ReGAIN architecture: pipeline from traffic ingestion to reasoning.}
\label{architecture}
\end{figure*}

Our proposed framework, ReGAIN, as illustrated in the Figure~\ref{architecture}, comprises four main components: (1) data ingestion and summarization, (2) semantic vectorization and knowledge base builder, (3) retrieval-augmented reasoning and generation, and (4) human-in-the-loop interaction.

\subsection{Data Ingestion and Summarization}

Network traffic telemetry originates from different sources, including log files, CSV anomaly annotations, and flow records. To enable uniform downstream processing, these inputs are normalized into a structured schema:
\begin{equation}
r_i = \{ ts_i, src_i, dst_i, p_i, proto_i, \ell_i \},
\end{equation}
where $ts_i$ is the timestamp, $src_i$ and $dst_i$ are the source and destination IPs, $p_i$ is the port, $proto_i$ is the protocol, and $\ell_i$ is the anomaly label. Each record is transformed into a natural-language summary:
\begin{equation}
s_i = f_{\text{sum}}(r_i),
\end{equation}
where $f_{\text{sum}}$ is a deterministic summarization function.  

For instance, a record such as  
\texttt{2024-08-15 10:05:23, 192.0.2.7, 203.0.113.5, icmp, label=DoS}  
is summarized as:  
\textit{“At 10:05:23 on August 15, 2024, host 192.0.2.7 sent an ICMP request to 203.0.113.5, flagged as a potential DoS anomaly.”}. 

This summarization serves several purposes. First, it reduces structured telemetry into an information-dense, human readable representation, preventing LLM context windows from being saturated by noise from raw logs. Second, it exposes network semantics (endpoints, protocol, timing, labels) in natural language, which improves the quality of embeddings and facilitates meaningful retrieval. Finally, providing concise, interpretable summaries ensures that when the ReGAIN cites supporting records as evidence, they are transparent and verifiable.

\subsection{Semantic Vectorization and Knowledge Base Builder}

Each natural-language summary is encoded into a $d$-dimensional embedding:
\begin{equation}
v_i = f_{\text{embed}}(s_i) \in \mathbb{R}^d,
\end{equation}
where $f_{\text{embed}}$ denotes a transformer-based embedding model. Each knowledge base entry is stored as:
\begin{equation}
e_i = (s_i, v_i, m_i),
\end{equation}
with $m_i$ containing structured metadata derived from the \emph{5-tuple} \((src\_IP, dst\_IP, src\_port, dst\_port, protocol)\), along with entry labels, and timestamps. To improve retrieval precision and context diversity, ReGAIN employs a \emph{multi-collection architecture} comprising three specialized vector databases: a telemetry collection containing enriched flow-level and packet-level summaries derived from PCAPs and log files, an anomaly collection capturing labeled or auto-detected attack instances with metadata, and a heuristic collection containing reference material such as detection heuristics and post incident annotations. Each collection is semantically indexed but remains logically isolated, allowing ReGAIN to retrieve context selectively or in parallel depending on query intent. For non-telemetry artifacts such as RFCs or incident tickets, the same representation applies: the document passage is treated as $s_i$, embedded into $v_i$, and tagged with relevant $m_i$ metadata.

\subsection{Retrieval-Augmented Reasoning and Generation}

When an analyst issues a query $q$, it is embedded into the same vector space as the corpus:
\begin{equation}
v_q = f_{\text{embed}}(q).
\end{equation}

Similarity is computed via cosine similarity:
\begin{equation}
\text{sim}(v_q, v_i) = \frac{v_q \cdot v_i}{\|v_q\| \, \|v_i\|}.
\end{equation}

To improve fidelity and prevent hallucination, we adopt a \emph{hierarchical retrieval strategy} that combines metadata-aware filtering with multi-stage semantic search. When a query is received, named entity and IP extraction automatically identify relevant metadata (e.g., destination IPs, protocols, ports, timestamps). These elements are used to construct a filter $\phi$ applied across all collections, narrowing the search to relevant flows or anomaly categories. Candidates satisfying $\phi$ are retrieved and ranked by semantic similarity:

\begin{equation}
R_\phi(q) = \{ e_i \in \mathcal{E}_\phi \mid \text{sim}(v_q, v_i) \geq \tau \}.
\end{equation}

where $\tau$ serves as the similarity threshold below which candidates are discarded.

To reduce redundancy, \emph{MMR} is applied to select a diverse subset that balances relevance and coverage, ensuring the LLM receives complementary evidence from telemetry, anomaly, and heuristic collections. These MMR-pruned candidates are then reranked hierarchically using a \emph{bi-encoder} followed by a \emph{cross-encoder}. The bi-encoder captures coarse semantic alignment, while the cross-encoder refines context sensitivity between the query and evidence pairs $(q, e_i)$.

Moreover, before generation, an \emph{abstention mechanism} implements a pre-generation quality gate to assess the coherence of retrieved evidence. If the number of high-confidence items $|R'_q|$ falls below a predefined threshold, the framework abstains from generating a response and returns undecidable, citing missing or inconsistent evidence. The top-$k$ results that pass quality checks are passed to the LLM:

\begin{equation}
y_q = f_{\text{LLM}}(q, R'_q),
\end{equation}

The output, $y_q$, follows a structured schema: a verdict (attack, no-attack, undecidable), the evidence and reasoning behind, and one or two recommended mitigations. If the similarity scores fall below a threshold $\tau$ or the evidence is inconsistent, the system abstains by outputting ``undecidable'' and listing missing context.

\subsection{Human-in-the-Loop Interaction}

The framework is designed as a decision-support tool, enabling network analysts to iteratively refine their investigations. Analysts can reformulate queries based on results:
\begin{equation}
q^{(t+1)} = g(q^{(t)}, y_q),
\end{equation}
where $g$ is an analyst-driven reformulation function.  

\textbf{Example:} An analyst may begin with a broad query (\textit{“Show anomalies involving 203.0.113.5”}), receive evidence of ICMP floods, and refine to a narrower one (\textit{“Compare with TCP SYN activity in the same interval”}). The framework maintains context across iterations, supporting forensic reasoning workflows that mimic real-world incident response.

\section{Experimental Setup} \label{experimental-setup}
This section describes the dataset, tools, and configurations used to implement and evaluate our framework. The software stack is summarized in Table~\ref{tab:tooling}.

\begin{table}[t]
\centering
\caption{Experimental environment and tooling}
\label{tab:tooling}
\begin{tabular}{@{}l p{6cm}@{}}
\hline
\textbf{Component} & \textbf{Configuration} \\ \hline
Data source & MAWILab v1.1 PCAPs (Jan 2022) \\
Parsing & Structured connection logs \\
Embeddings & all-MiniLM-L6-v2 (384-D) \\
Cross-encoder & cross-encoder/ms-marco-MiniLM-L-6-v2\\
Vector store & ChromaDB with three persistent collections \\
Orchestration & LangChain framework \\
LLM & GPT-4.1-nano, temperature = 0 \\
Similarity threshold & $\tau = 0.3$ \\
MMR parameters & $k=3$--$6$, fetch\_k=$3k$\\ \hline
\end{tabular}
\end{table}

\paragraph{Dataset and Knowledge Base} 
We use the MAWILab v1.1 network traces~\cite{mawilab}. For each day, MAWILab provides (i) PCAPs and (ii) structured anomaly CSVs with two complementary labels, comprising \emph{heuristic} (signature/flag/port/type--code driven) and \emph{taxonomy} (behavioral categories such as DoS, scans, tunneling). Each row includes the 5-tuple (where applicable), heuristic/taxonomy codes, severity (anomalous, suspicious, notice, benign), and identifiers. In this study, we focus on analyzing ICMP and TCP network activities, specifically ping flood and SYN flood attacks. We selected raw MAWILab PCAPs captured on January 1, 9, and 10, 2022. These captures reflect contemporary Internet conditions and protocol distributions, and also retain the detailed anomaly annotations required for reproducible evaluation. We store the embeddings and metadata in ChromaDB~\cite{chroma}, an open-source vector database designed for high-dimensional similarity search. 

\begin{figure}[htbp]
    \centering
    \includegraphics[width=\columnwidth]{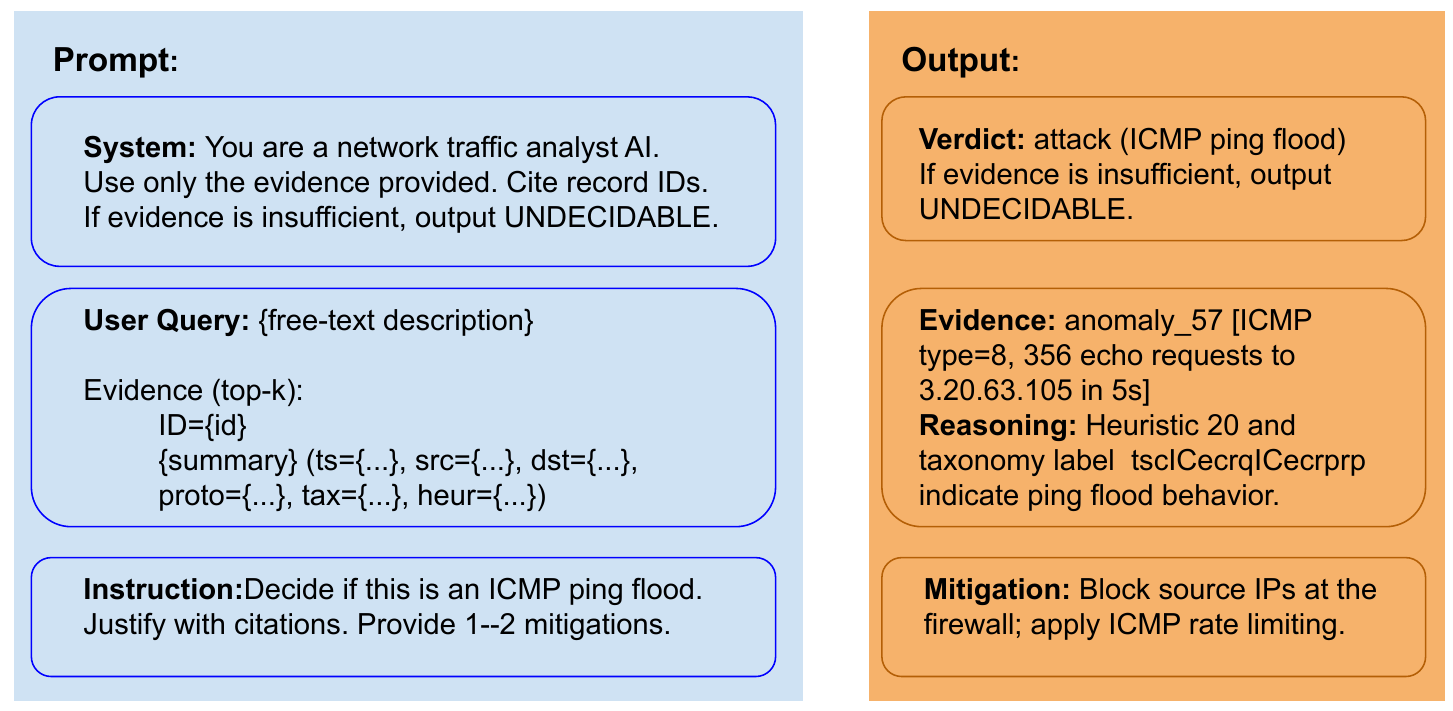}
    \caption{An abridged prompt and output of the system.}
    \label{fig:prompt-output}
\end{figure}

\paragraph{Prompt Structure}
The prompt template controls how retrieved evidence is presented to the language model. Our design goal was to balance three requirements: (i) grounding the model’s reasoning in verifiable evidence, (ii) enforcing a consistent and auditable output format, and (iii) ensuring actionable recommendations for operators. The system instruction requires the model to cite retrieved record IDs or heuristic codes in its reasoning. If the retrieval context is insufficient, the model is instructed to output the keyword \textit{undecidable} and list the missing evidence.  The response schema follows a three-part structure, incorporating an alert summary describing the detected activity, a justification citing retrieved evidence, and one or two concise mitigation steps. The prompt explicitly instructs the model to provide assertive, confident assessments and avoid hedging language (e.g., ``might'', ``possibly''), which makes outputs more actionable rather than tentative. In the deployed environment, the prompt and model output are displayed through a lightweight command-line interface (CLI). This interface enables analysts to inspect retrieved evidence, view the structured LLM response, and iteratively refine their queries. Figure~\ref{fig:prompt-output} demonstrates an abridged prompt and output of the framework. 

\paragraph{Inference Parameters}
To ensure the comparability of results, we use a uniform instruction block across all runs. We also employ deterministic decoding (temperature $\approx 0$) to minimize variability in the LLM outputs. We cap the retrieved evidence at $k \in \{3,5\}$: a smaller $k$ reduces context dilution and enforces concise reasoning, while a larger $k$ provides additional corroborating records. This range was selected empirically as a balance between precision (avoiding irrelevant context) and recall (ensuring sufficient evidence is available).

\section{Evaluation} \label{evaluation}

We evaluated the ReGAIN framework using two complementary methodologies across two attack scenarios, TCP SYN floods and ICMP ping floods. The first method involved an automated comparison against ground-truth annotations, and the second one incorporated manual expert judgment to assess the framework’s performance independently. Table~\ref{tab:attack-metrics} summarizes key performance metrics for SYN and Ping flood attacks under both ground-truth and expert labels. Performance was evaluated using standard metrics. Accuracy was computed as 
\( \frac{TP + TN}{TP + TN + FP + FN} \); Precision as 
\( \frac{TP}{TP + FP} \); Recall as 
\( \frac{TP}{TP + FN} \); and the F1 score as 
\( 2 \times \frac{\text{Precision} \times \text{Recall}}{\text{Precision} + \text{Recall}} \).

To generate ground truth labels, SYN flood attacks were identified using TCP connection states in the logs. Connections marked as \texttt{S0} (SYN sent, no reply) indicate incomplete handshakes typical of SYN floods, and those marked as (\texttt{SH}, \texttt{SF}, \texttt{RSTR}, \texttt{RSTO}, \texttt{OTH}) reflect normal or benign traffic. For ping flood attacks, a sliding-window detector flagged logs with ten or more ICMP echo requests (type 8) from the same source to the same destination within a 20-second window as attacks, capturing both one-to-many and many-to-one flooding behavior. These automated labels were cross-validated with MAWILab’s heuristic-20 anomalies to improve labeling accuracy.

To establish an independent validation baseline, a subset of connection logs underwent blind expert adjudication based on traffic characteristics and domain knowledge. Experts analyzed Zeek logs using known attack signatures. For SYN flood detection, experts assessed TCP connection states, labeling events with high volumes of incomplete handshakes (\texttt{S0}) from specific IPs as attacks. They differentiated attacks from benign failures by considering connection rate, IP diversity, timing, and correlation with MAWILab anomalies. For ping flood detection, ICMP traffic was examined based on type codes, IP pairs, and timing. Events were labeled as attacks if  MAWILab listed the IPs with heuristic code 20, or ten or more ICMP echo requests from a single source to a destination occurred in a short time frame. Logs with 5–9 requests were flagged for further review, while those with fewer than five were deemed benign.

\subsection{SYN Flood Attack}

\begin{figure}[!t]
    \centering
    \subfloat[Confusion Matrix \label{fig:syn-gt-cm}]{
        \includegraphics[width=0.45\linewidth]{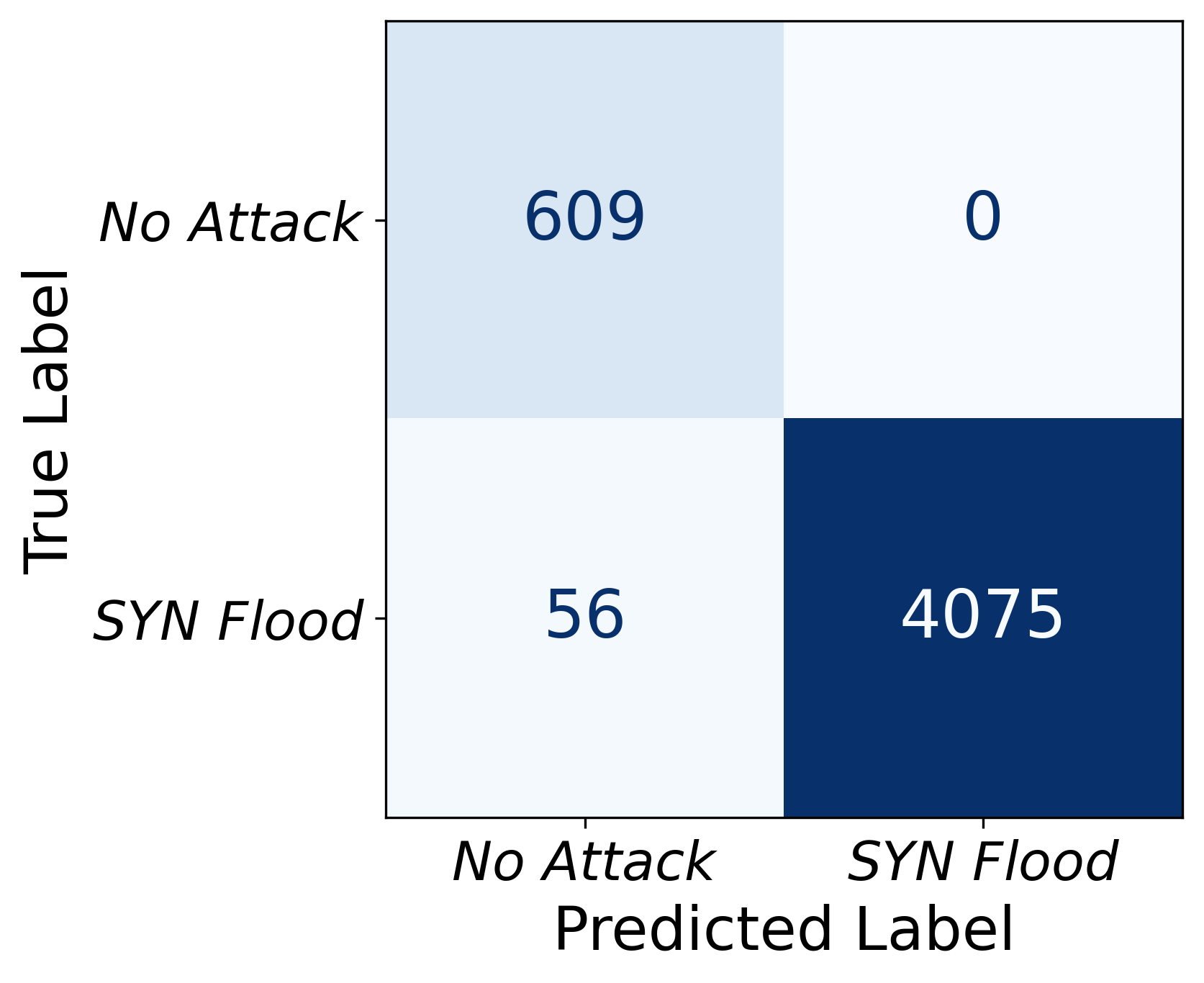}
    }\hfill
    \subfloat[ROC Curve \label{fig:syn-gt-auc}]{
        \includegraphics[width=0.45\linewidth]{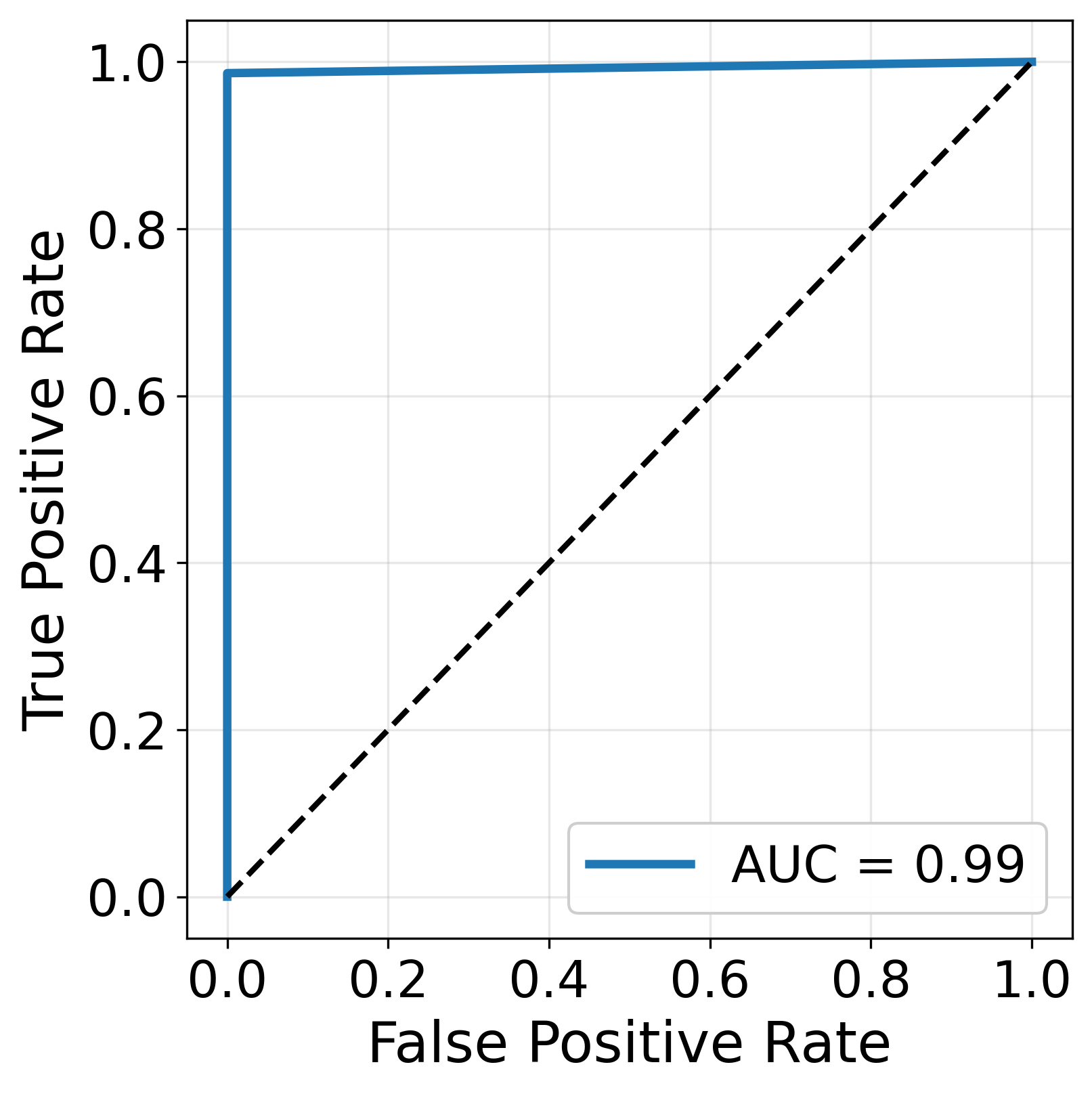}
    }
    \caption{SYN Flood Evaluation Against Ground Truth: (a) Confusion matrix, (b) ROC curve.}
    \label{fig:syn-gt}
\end{figure}

\paragraph{Results Against Ground Truth} As illustrated in Figure \ref{fig:syn-gt-cm}, the confusion matrix reveals a highly favorable distribution with 4,075 true positives, 609 true negatives, zero false positives, and only 56 false negatives. This corresponds to an overall accuracy of 98.82\%, precision of 100.00\%, recall of 98.64\%, and an F1 score of 99.32\%. The ROC curve in Figure \ref{fig:syn-gt-auc} further illustrates this discriminative capability, yielding an AUC of 0.99. The perfect precision indicates that when ReGAIN identifies SYN flood activity, it does so with complete certainty, while the near-perfect recall demonstrates that very few actual attacks escape detection.

\begin{figure}[!t]
    \centering
    \subfloat[Confusion Matrix \label{fig:syn-ex-cm}]{
        \includegraphics[width=0.45\linewidth]{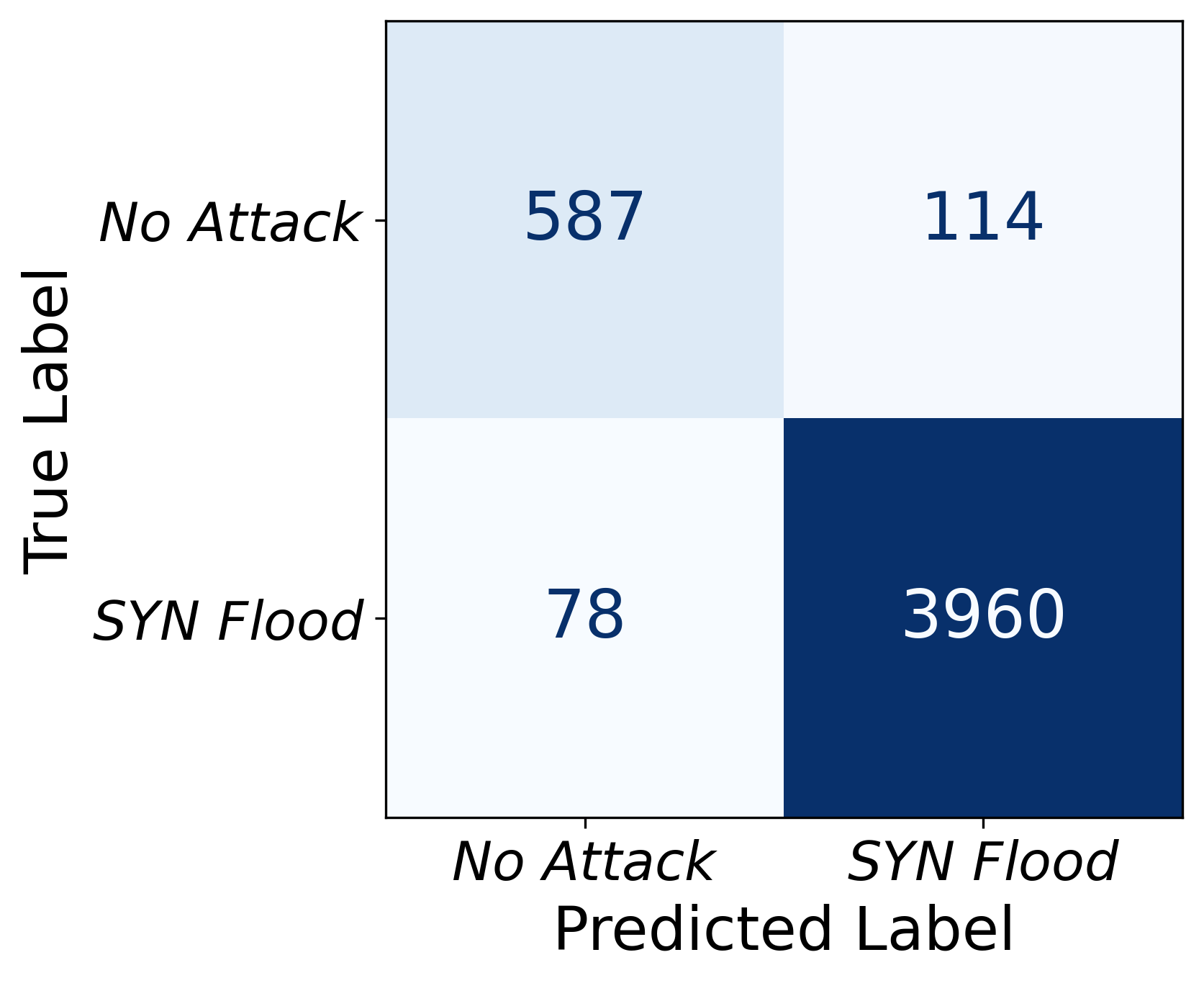}
    }\hfill
    \subfloat[ROC Curve \label{fig:syn-ex-auc}]{
        \includegraphics[width=0.45\linewidth]{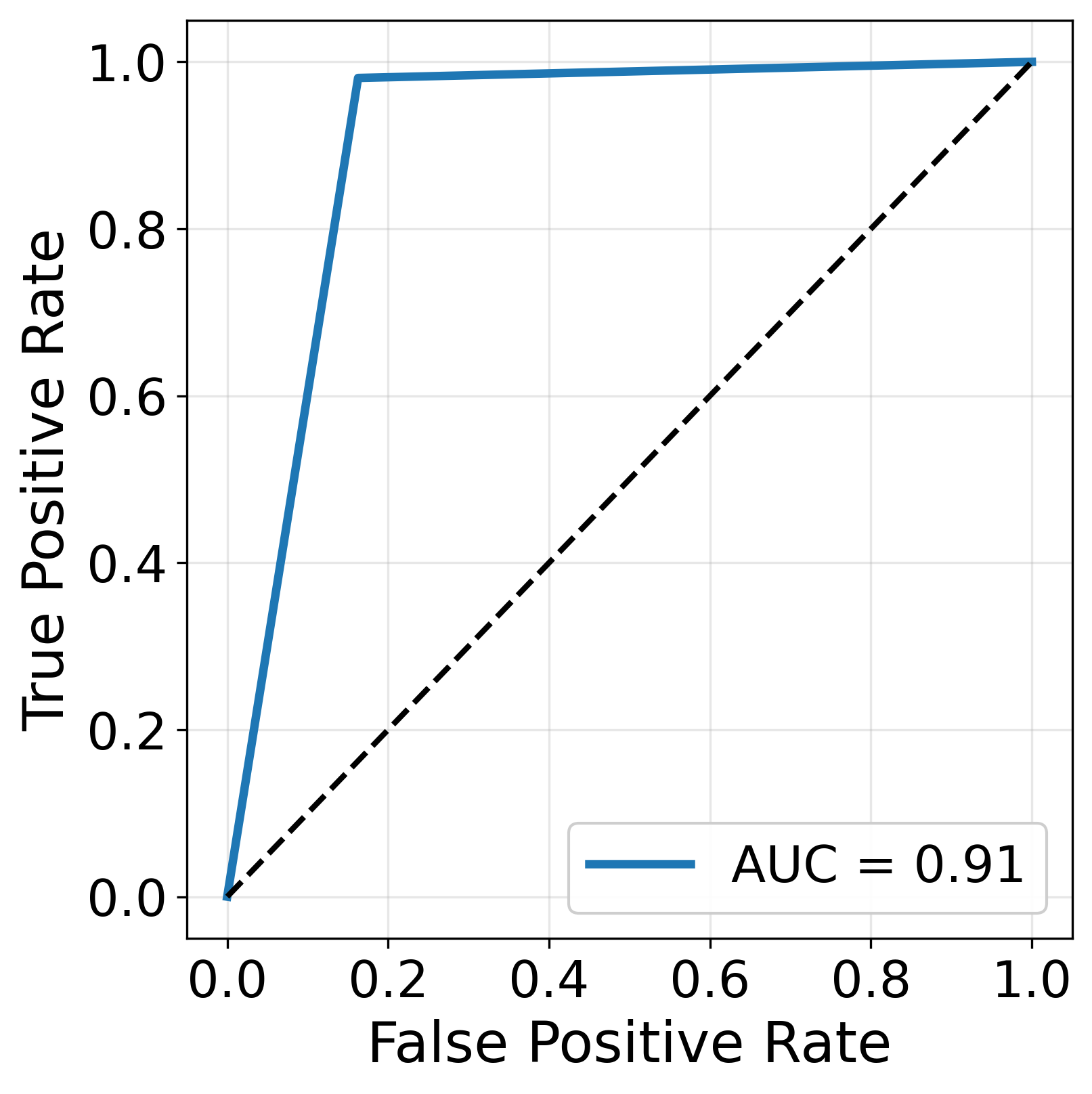}
    }
    \caption{SYN Flood Evaluation Against Expert Labels: (a) Confusion matrix, (b) ROC curve.}
    \label{fig:syn-ex}
\end{figure}

\paragraph{Results Against Expert Labels} Expert evaluation of the SYN flood attack revealed a different pattern of refinement, where the confusion matrix (Figure \ref{fig:syn-ex-cm}) shows 3,960 true positives, 587 true negatives, 114 false positives, and 78 false negatives. This represents a shift from the ground-truth evaluation, showing that although precision remained exceptionally high at 97.20\%, recall decreased slightly to 98.07\%, resulting in an overall accuracy of 95.95\% and an F1 score of 97.63\%. The ROC curve in Figure \ref{fig:syn-ex-auc} exhibits continued strong discriminative performance with an AUC of 0.98. The emergence of false positives and false negatives under expert review suggests that some automated ground-truth labels may have been overly permissive, and that certain edge cases, such as partial SYN floods or rate-limited attack attempts, require human judgment to classify accurately.

\subsection{Ping Flood Attack}

\paragraph{Results Against Ground Truth}

\begin{figure}[!t]
    \centering
    \subfloat[Confusion Matrix \label{fig:png-gt-cm}]{
        \includegraphics[width=0.45\linewidth]{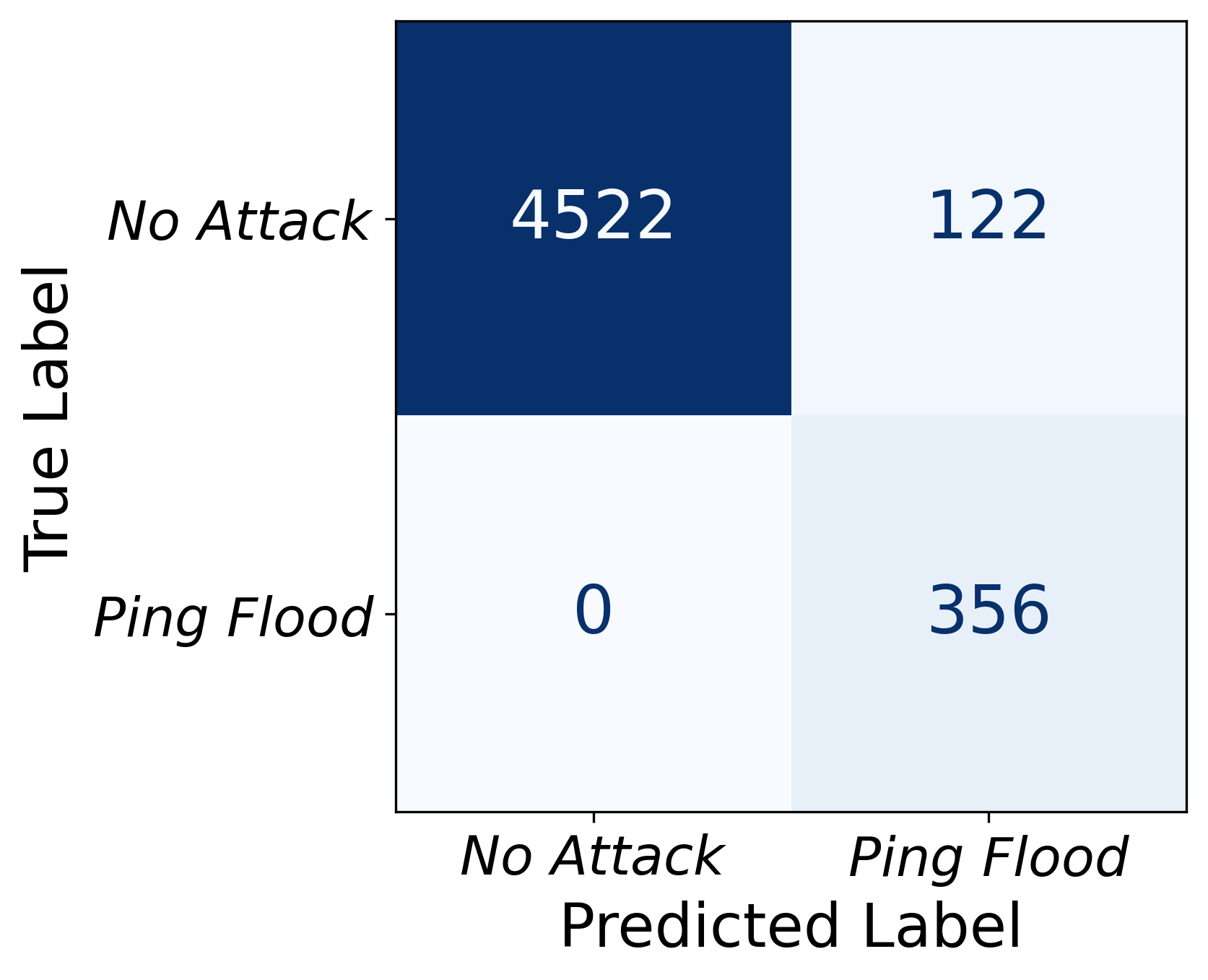}
    }\hfill
    \subfloat[ROC Curve \label{fig:png-gt-auc}]{
        \includegraphics[width=0.45\linewidth]{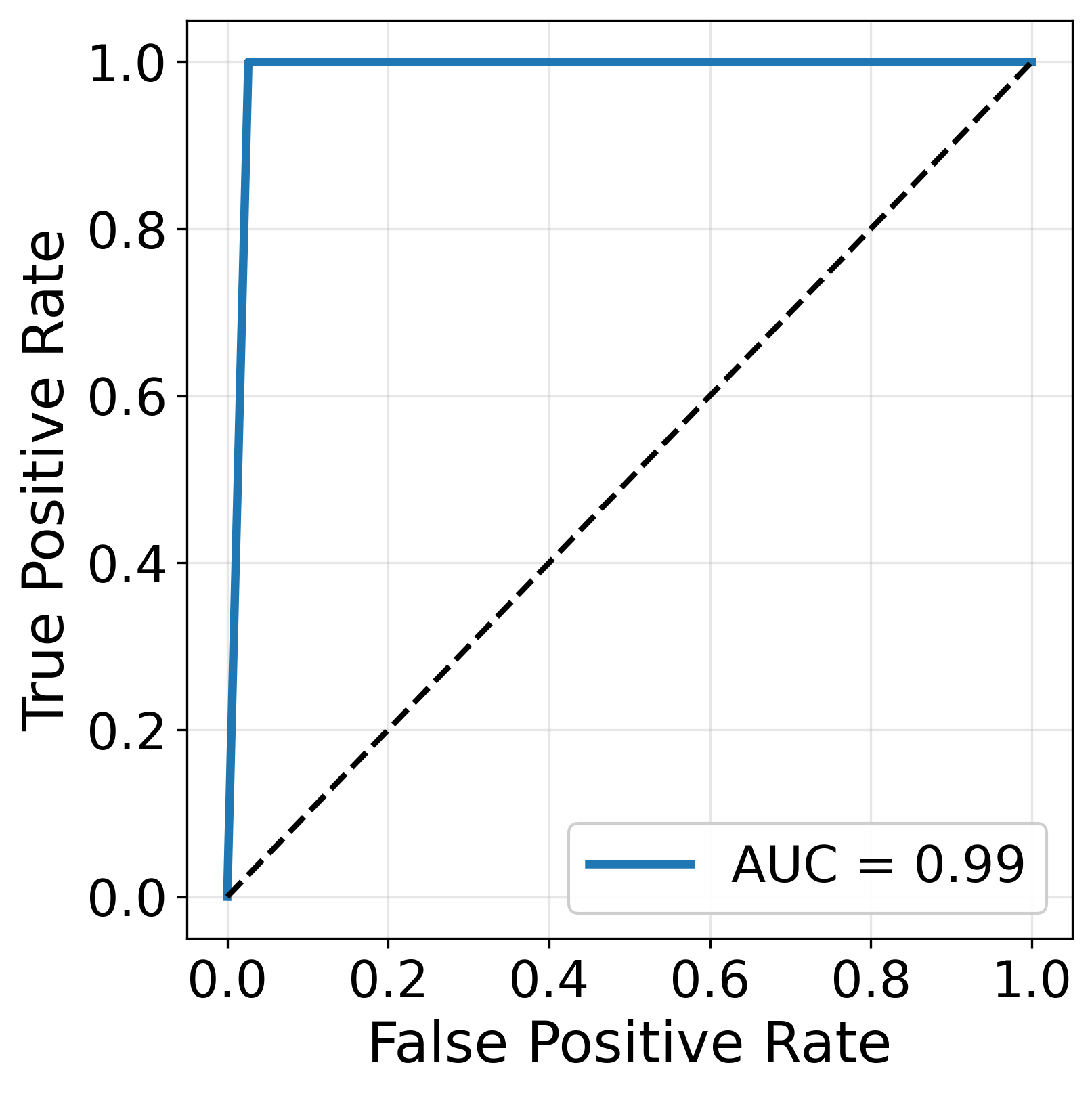}
    }
    \caption{Ping Flood Evaluation Against Ground Truth: (a) Confusion matrix, (b) ROC curve.}
    \label{fig:png-flood-gt}
\end{figure}

For the ping flood attack, the system achieved perfect recall (100.00\%) while identifying all 356 true attack instances with zero false negatives, as shown in Figure \ref{fig:png-gt-cm}. This perfect sensitivity came at the cost of precision, which measured 74.48\% due to 122 false positives, benign ICMP traffic misclassified as attacks. The overall accuracy reached 97.56\% with an F1 score of 85.37\%. The ROC curve in Figure \ref{fig:png-gt-auc} demonstrates strong discriminative capability with an AUC of 0.99, indicating near-perfect class separability despite the precision trade-off. These results suggest that although the system maintains exceptional recall for ping flood attacks, its precision is affected by the similarity between benign diagnostic ICMP activity and actual flood patterns.

\paragraph{Results Against Expert Labels}

\begin{figure}[!t]
    \centering
    \subfloat[Confusion Matrix\label{fig:png-ex-cm}]{
        \includegraphics[width=0.45\linewidth]{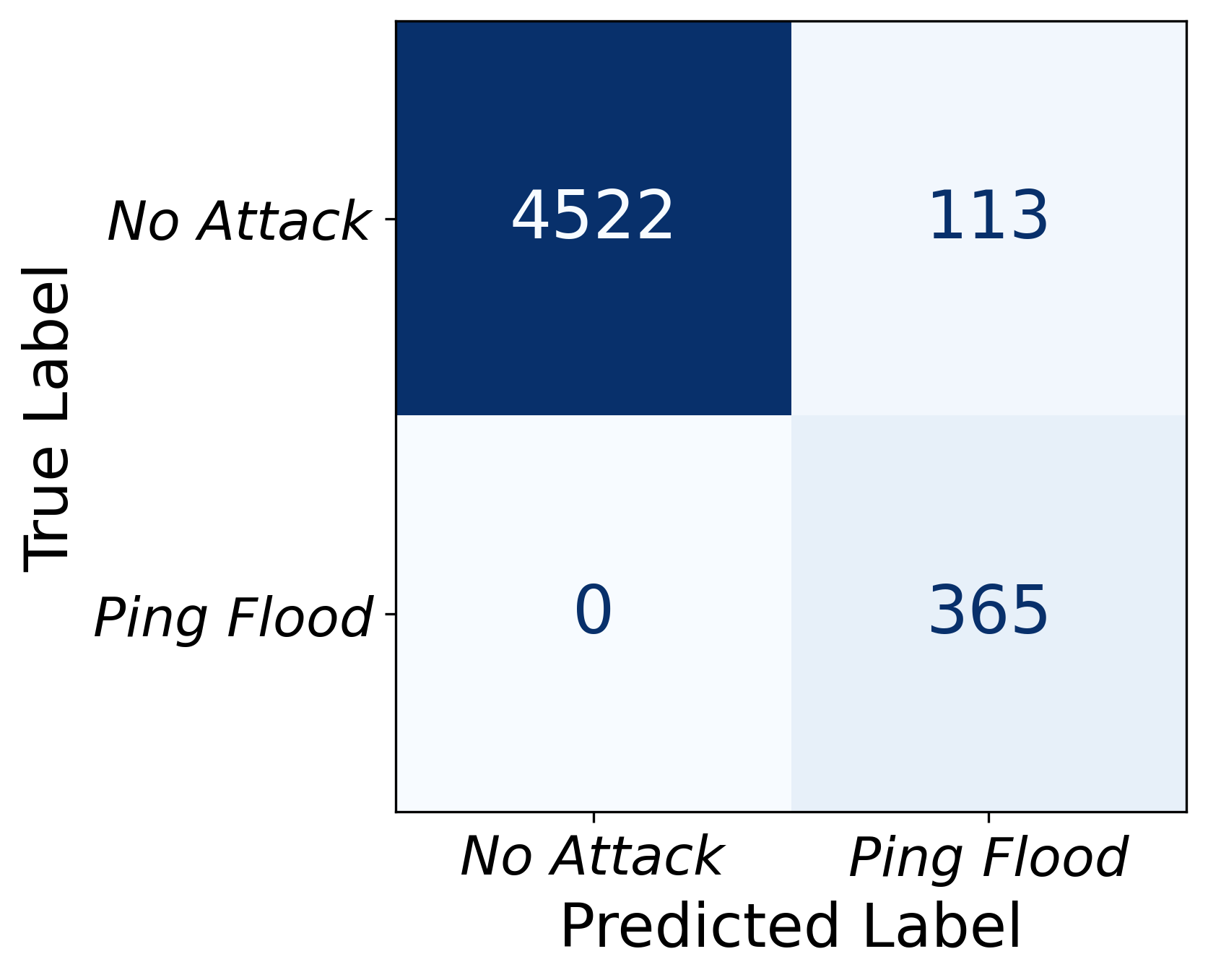}
    }\hfill
    \subfloat[ROC Curve\label{fig:png-ex-auc}]{
        \includegraphics[width=0.45\linewidth]{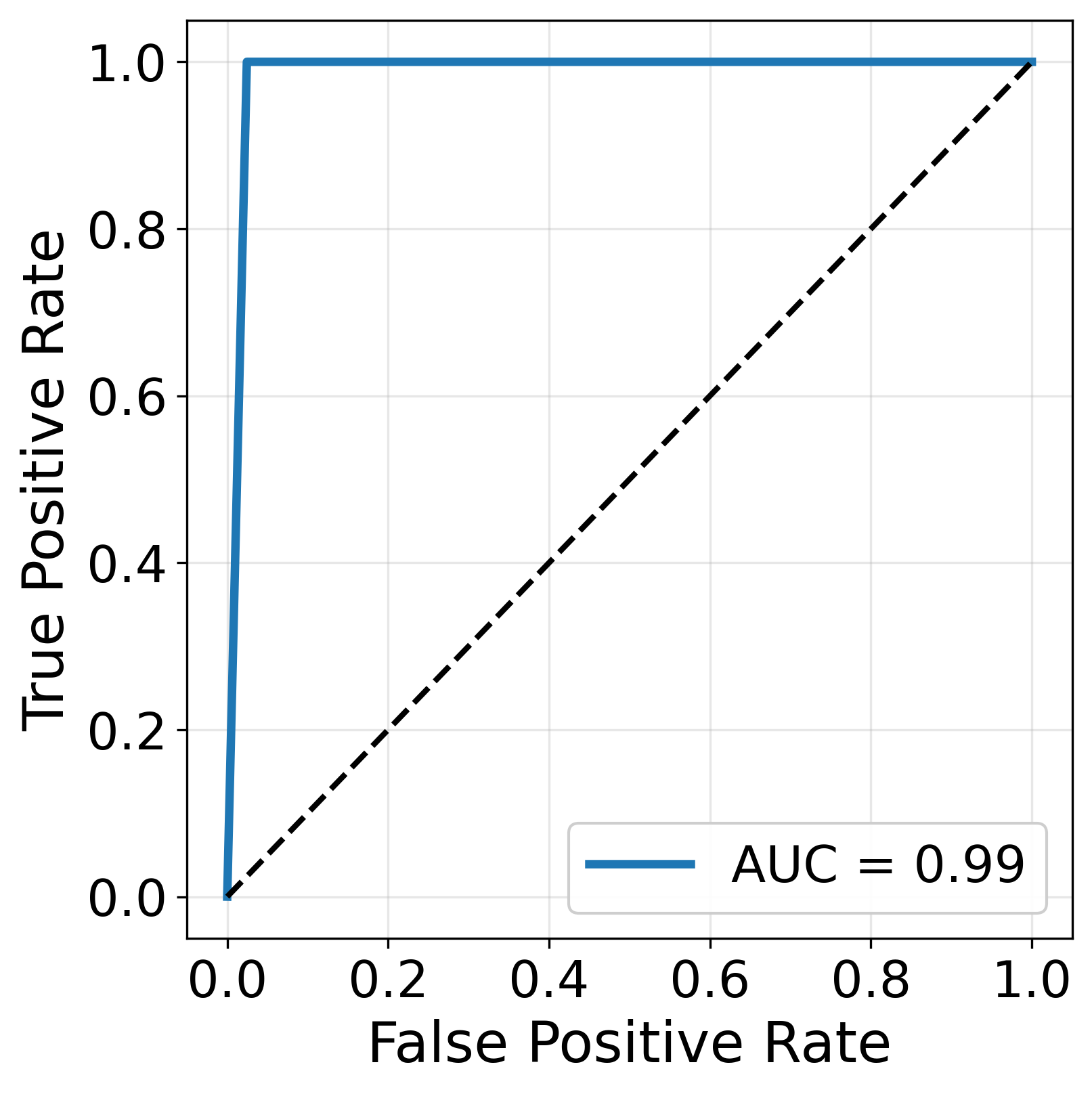}
    }
    \caption{Ping Flood Evaluation Against Expert Labels: (a) Confusion matrix, (b) ROC curve.}
    \label{fig:png-flood-ex}
\end{figure}

In expert evaluation, the ping flood attack maintained perfect recall (100.00\%), correctly identifying all 365 validated attack instances without any false negatives, as shown in Figure \ref{fig:png-ex-cm}. Compared to the ground-truth evaluation, the expert labels identified nine additional true positive cases (365 vs. 356), representing instances where the framework correctly detected attacks that were missing or underrepresented in the original automated annotations. The number of false positives decreased modestly from 122 to 113, which increased the precision to 76.36\% and improved overall accuracy to 97.74\%. The F1 score increased to 86.60\%, and the ROC curve in Figure \ref{fig:png-ex-auc} maintained an AUC of 0.99. 

The reduced precision in our results shows that, despite ReGAIN's perfect recall in both evaluations, it has a tendency to overclassify certain benign ICMP activities as attacks. Most false positives originated from short-lived or diagnostic ICMP bursts, such as network reachability checks, latency probes, or automated monitoring tasks, that temporarily showed traffic patterns similar to genuine ping floods. Since MAWILab annotations do not always distinguish between benign high-frequency ICMP traffic and attack-induced floods, the framework conservatively labeled these cases as anomalous.

\begin{table}[t]
\centering
\small 
\caption{Performance Metrics for SYN and Ping Flood Attacks}
\label{tab:attack-metrics}
\begin{tabular}{@{}lccccc@{}}
\hline
\textbf{Attack} & \textbf{Acc.} & \textbf{Prec.} & \textbf{Recall} & \textbf{F1} & \textbf{AUC} \\ \hline
SYN (GT)   & 98.82\% & 100.00\% & 98.64\% & 99.32\% & 0.99 \\
SYN (Expert) & 95.95\% & 97.20\% & 98.07\% & 97.63\% & 0.98 \\
Ping (GT)  & 97.56\% & 74.48\% & 100.00\% & 85.37\% & 0.99 \\
Ping (Expert) & 97.74\% & 76.36\% & 100.00\% & 86.60\% & 0.99 \\ \hline
\end{tabular}
\end{table}

\subsection{Comparison with Baseline Methods}

To evaluate the effectiveness of ReGAIN, we conducted a comparative analysis against five baseline detection approaches: (1) traditional rule-based intrusion detection employing Snort-style threshold heuristics, (2) SVM with Radial Basis Function kernels, (3) Random Forest ensemble classifiers configured with 100 decision trees, (4) one-dimensional Convolutional Neural Networks (CNN), and (5) two-layer Long Short-Term Memory (LSTM) networks. All models use 30 numerical features extracted from Zeek conn.log files (protocol ratios, connection states, temporal stats, byte/packet volumes, IP/port diversity) and these features are normalized using StandardScaler (z-score). Also, all models were trained and evaluated on identical partitions of the MAWILab dataset to ensure methodological consistency, with performance metrics calculated using ground-truth labels.

\begin{figure}[!t]
    \centering
    \subfloat[SYN Flood Comparison\label{fig:acc-syn}]{
        \includegraphics[width=0.9\linewidth]{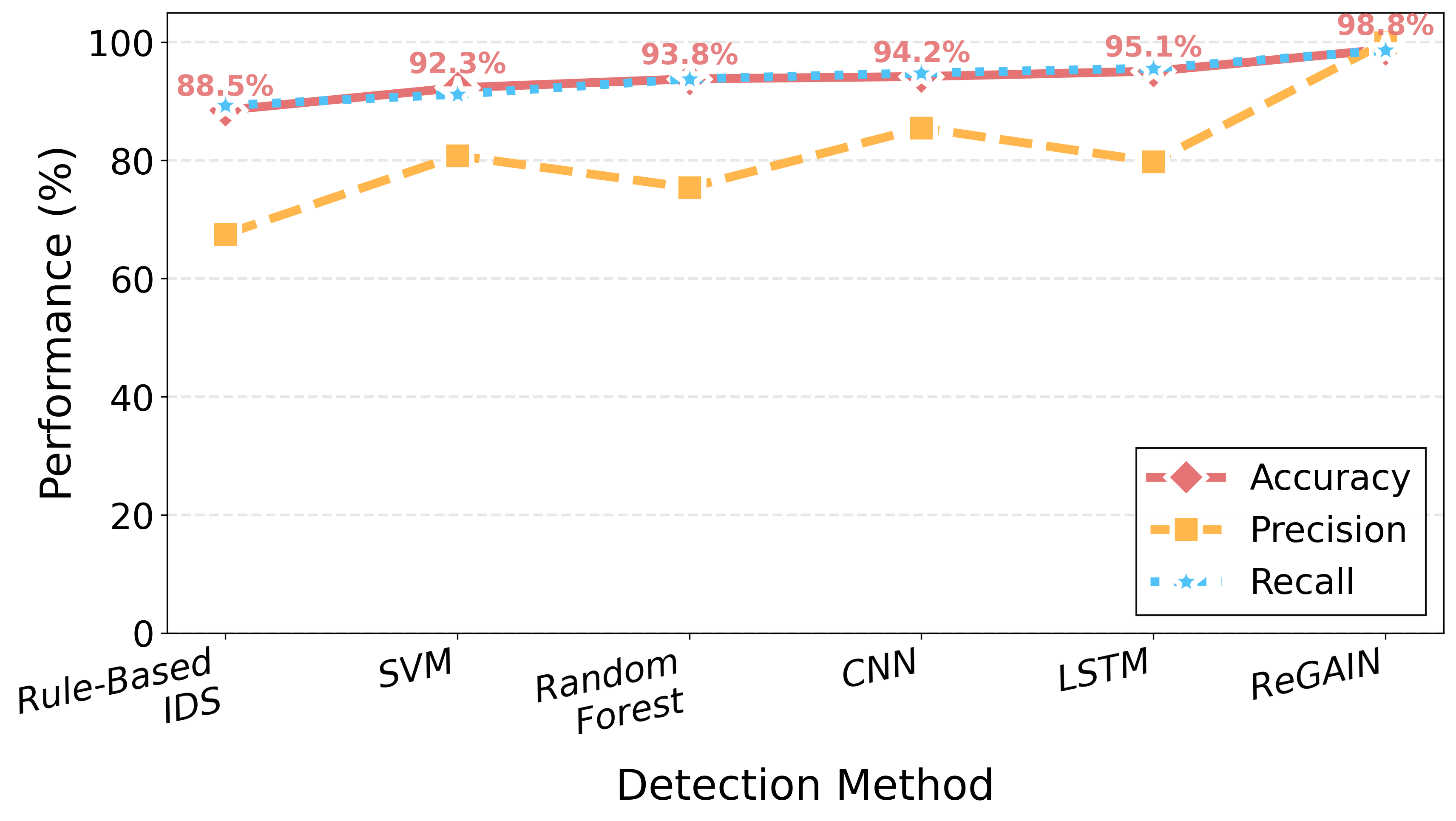}
    }
    \vspace{1em}
    \subfloat[Ping Flood Comparison\label{fig:acc-ping}]{
        \includegraphics[width=0.9\linewidth]{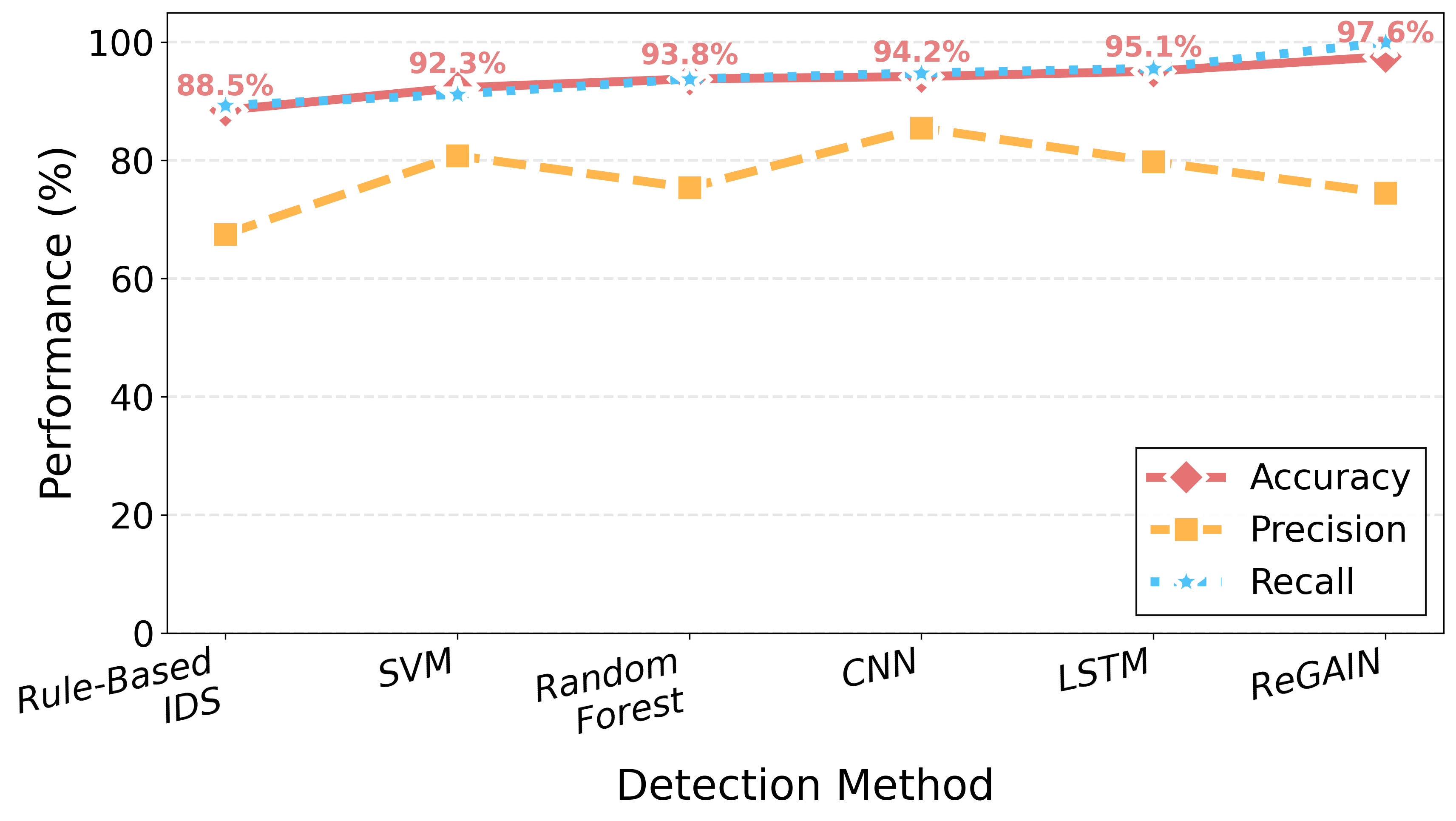}
    }
    \caption{Performance comparison of ReGAIN against baseline detectors for Ping Flood and SYN Flood attacks.}
    \label{fig:accuracy-comparison}
\end{figure}

Figure~\ref{fig:accuracy-comparison} presents a comparative analysis of ReGAIN and the baselines. For SYN Flood attack (Figure~\ref{fig:acc-syn}), compared to the best-performing baseline (LSTM), ReGAIN improves accuracy by 3.7 percentage points, precision by 14.5 points, and recall by 3.8 points. This balanced performance profile demonstrates ReGAIN's ability to maintain high detection sensitivity and simultaneously minimize false positives, which is a challenging combination for conventional machine learning approaches.

For the Ping Flood attack (Figure \ref{fig:acc-ping}), ReGAIN achieves superior overall performance, surpassing the strongest baseline (LSTM). Most notably, ReGAIN achieves perfect recall (100.0\% vs. 95.6\% for LSTM), ensuring zero false negatives at the cost of reduced precision (74.5\%). This precision-recall trade-off reflects a deliberate design choice that prioritizes detection sensitivity, i.e., a critical requirement in network security applications where failing to detect attacks carries substantially higher risk than investigating false alarms.

Beyond quantitative performance improvements, ReGAIN offers critical qualitative advantages that distinguish it from traditional detection systems. ReGAIN generates human-readable natural language explanations for each detection decision. This interpretability addresses a fundamental limitation in network security operations, where security analysts require actionable insights rather than opaque verdicts. Furthermore, ReGAIN's conversational interface enables interactive refinement of detection criteria and iterative querying of network behavior, facilitating collaborative human-AI investigation workflows that are infeasible with static classification models. These capabilities transform network traffic analysis from a passive alerting mechanism into an interactive analytical tool, supporting both automated detection and human-guided threat hunting activities.

\section{Conclusion and Future Directions} \label{conclusion}

This work presented ReGAIN, a novel framework that integrates network traffic summarization, semantic search, and LLM-driven reasoning to support transparent and accurate traffic analysis. By transforming raw network data into descriptive, embedded summaries stored in a multi-collection vector database, ReGAIN enables efficient, evidence-grounded retrieval of relevant historical patterns. The system’s hierarchical retrieval and re-ranking mechanisms, combined with metadata filtering and an abstention strategy, help mitigate hallucinations and ensure interpretability. Evaluations of ICMP ping flood and TCP SYN flood scenarios, containing 10,000 labeled instances, demonstrated ReGAIN’s strong performance, with high accuracy (95.95–98.82\%) and near-perfect recall (98.64–100\%), outperforming traditional rule-based and learning-based baselines. These results show the promise of LLM-driven reasoning and retrieval augmentation for network traffic analysis. 

There are several limitations and areas for improvement that we aim to address in future work. Although ReGAIN demonstrates strong performance, it relies on GPT-4-class models accessed via remote APIs and utilizes lightweight 384-dimensional embeddings designed to favor reasoning quality over computational throughput. As a result, inference latency is constrained by API communication, making ReGAIN more appropriate for retrospective analysis and forensic investigations rather than real-time monitoring. To reduce latency and enhance data privacy, future deployments can incorporate on-premise models such as LLaMA-3-8B, Mistral-7B, or Phi-3, and local embedding models to enable fully offline operation in air-gapped or sensitive environments. 

We also note a precision gap (approximately 74–76\%) in detecting ping flood attacks, primarily due to benign ICMP activity that can resemble attack patterns. To improve detection accuracy, we plan to consider dynamic similarity thresholds based on protocol type and traffic volume, as well as temporal and rate-based filters to differentiate short bursts from sustained attacks. In addition, future work will focus on evaluating the clarity and effectiveness of the generated natural language outputs, ensuring that summaries and explanations are not only accurate but also easily interpretable by analysts. These directions will strengthen ReGAIN as a practical, lightweight, and explainable framework for modern network traffic analysis.

\section*{Acknowledgment}
This work is partially supported by NSF grants 2113945, 2200538, 2416992, and 2230610 at NC A\&T SU. Any opinions, findings, and conclusions or recommendations expressed in this material are those of the author(s) and do not necessarily reflect the views of the funding agency.

\end{document}